\documentclass[runningheads]{llncs}

\usepackage{eccv}

\usepackage{eccvabbrv}

\usepackage{graphicx}
\usepackage{booktabs}

\usepackage[accsupp]{axessibility}  

\usepackage{hyperref}

\hypersetup{
    colorlinks,
    citecolor=eccvblue,
    filecolor=black,
    linkcolor=black,
    urlcolor=eccvblue
}

\usepackage{orcidlink}

\usepackage{threeparttable} 
\usepackage{pifont} 
\usepackage{float}

\usepackage{marvosym} 

\newcommand{\datasetname}{{\sc SynHead100 }}

\begin{document}

\title{Head360: Learning a Parametric 3D Full-Head for Free-View Synthesis in 360$^\circ$\\} 

\titlerunning{Head360}

\author{Yuxiao He$^{1}$, Yiyu Zhuang$^{1}$, Yanwen Wang$^{1}$, Yao Yao$^{1}$, Siyu Zhu$^2$, \\
Xiaoyu Li$^3$, Qi Zhang$^3$, Xun Cao$^{1}$, Hao Zhu$^{1}$$\textsuperscript{\Letter}$}

\authorrunning{Y.~He et al.}

\institute{$^1$ State Key Laboratory for Novel Software Technology, Nanjing University, China\\
$^2$ Fudan University, Shanghai, China \quad  $^3$ Tencent AI Lab, Shenzhen, China
}

\maketitle

\begin{abstract}

Creating a $360^\circ$ parametric model of a human head is a very challenging task. While recent advancements have demonstrated the efficacy of leveraging synthetic data for building such parametric head models, their performance remains inadequate in crucial areas such as expression-driven animation, hairstyle editing, and text-based modifications. In this paper, we build a dataset of artist-designed high-fidelity human heads and propose to create a novel parametric 360-degree renderable parametric head model from it. Our scheme decouples the facial motion/shape and facial appearance, which are represented by a classic parametric 3D mesh model and an attached neural texture, respectively. We further propose a training method for decompositing hairstyle and facial appearance, allowing free-swapping of the hairstyle. A novel inversion fitting method is presented based on single image input with high generalization and fidelity. To the best of our knowledge, our model is the first parametric 3D full-head that achieves $360^\circ$ free-view synthesis, image-based fitting, appearance editing, and animation within a single model. Experiments show that facial motions and appearances are well disentangled in the parametric space, leading to SOTA performance in rendering and animating quality. The code and \datasetname dataset are released in \url{https://nju-3dv.github.io/projects/Head360}.

\end{abstract}

\section{Introduction}
\label{sec:intro}

3D head modeling has been a longstanding and hot research topic in computer vision and graphics and is essential to many human-related downstream tasks. Generating a high-fidelity head model involves 3D surface reconstruction, material modeling, hair designing, rigging, etc, which is very complicated. To solve these challenges, classic pipelines rely on expensive systems (like light stage~\cite{debevec2012light}) and the inevitable large amount of manpower (like hair design and rigging correction) to produce a high-fidelity head model. In recent years, the emergence of neural rendering and neural texture techniques has made it possible to implement this pipeline end-to-end with a sophisticated neural network, thus fully automated through a data-driven approach.

Overall, the head modeling methods based on neural rendering can be divided into two technical routings. The first routing leverages massive 2D face image sets for training and achieves highly realistic rendering, motion retargeting, and identity inversion~\cite{chan2022efficient, sun2023next3d}. The downsides are obvious, that is, it can only achieve high-quality renderings at a small angle range on the frontal side ($<\pm30^\circ$), while larger poses will cause serious image quality degradation. The second routing is to learn neural heads from 3D models, which are scanned real faces~\cite{yang2020facescape, dai2020statistical, wang2022faceverse, zhu2023facescape} or models designed by artists~\cite{wang2023rodin, wood2021fake}. These methods can synthesize $360^\circ$ high-quality renderings. However, the performance of motion animation and identity inversion is worse since too few 3D models are available for training. In summary, existing parametric 3D heads inevitably suffer from limited renderable angles (those trained on massive 2D images) or poor rigging/identity inversion (those trained on limited 3D head models).

\begin{figure}[t]
    \centering
    \includegraphics[width=1.0\linewidth]{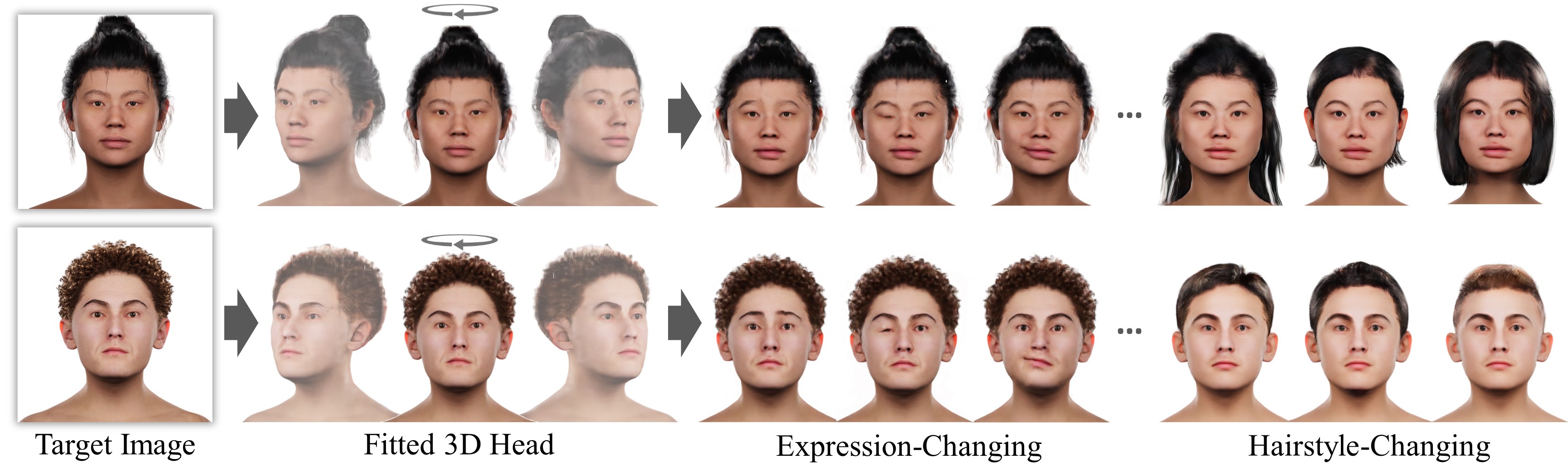}
    \caption{
    \textbf{Overview.} Our model is the first $360^{\circ}$-renderable parametric 3D head with hair that supports image-based fitting and animation simultaneously.
    }
   
    \label{fig:title}
\end{figure}

In this paper, we go further along the routine of modeling human heads from artist-designed 3D models.
Since previous artist-designed datasets are not publicly available, we produce and schedule to open-source a high-fidelity 3D head dataset of 100 identities and 52 standard defined expressions per identity. To the best of our knowledge, this is the first publicly available large-scale high-fidelity 3D human head dataset. Based on the dataset, a hybrid representation is leveraged to model the 4D head with hair, adopting mesh and its parametric model to represent 3D shape/motion while adopting neural texture to represent the appearance. The shape and appearance are then transformed into a radiance field and are rendered via the volume renderer. We split the radiation field into two parts, representing the head and the hair, respectively, so that the hairstyle can be switched freely. With this framework, we train a parametric head model with hair that is $360^\circ$ renderable.

Based on this parametric model, we further extend its application for image-based fitting and appearance editing. A novel fitting method is designed to fit the shape and appearance of the whole human head based on a single image input.  The generated or fitted head can be stylized according to a text prompt by introducing the prior from the large vision-language model. We are surprised that the fitted or edited head can still be animated with standard blendshapes parameter streams in high quality. We consider the reason because the framework of `parametric 3D mesh $+$ neural texture' effectively disentangles the head motion and appearance. Experiments show that our model can generate high-quality 3D head models that outperform previous parametric heads and are capable of fitting, animating, and text-based editing.

Our contribution can be summarized as:

\begin{itemize}
    \item We propose a \textit{parametric 3D mesh $+$ neural texture} for parameterizing 3D heads, which has proved effective in disentangling head motions and appearances.

    \item A novel framework is proposed to detach hair and head, leading to the capability of hairstyle swapping and stronger fitting performance.
    \item Image-based fitting and text-based stylizing schemes are proposed specialized for our parametric head fitting and editing.
    \item A high-fidelity artist-designed 3D head dataset consisting of 100 identities, each with 52 standard expressions, is created and will be released upon publication.
\end{itemize}

\section{Related Work}
\label{sec:related}

\noindent\textbf{3D Morphable Model (3DMM): }
3DMM is defined as a statistical model that transforms the 3D shape and texture of the heads into a vector space representation\cite{blanz1999morphable}. The parametric transformation is learned from a large set of heads in diverse shapes, appearances, and expressions, which are represented by registered 3D polygonal mesh models. According to the properties of parametric transformation, 3DMM can be further divided into linear models and nonlinear models. For a linear 3DMM, Principal Component Analysis (PCA)~\cite{pearson1901liii} is commonly used to learn the linear mapping from the 3D vertices and texture space of the head model to a low-dimensional space~\cite{vlasic2005face, cao2013facewarehouse, li2017learning, jiang2019disentangled, yang2020facescape, zhu2023facescape, JNR}. Linear 3DMM conversion calculation is straightforward and is widely used in face alignment, face reconstruction, and other tasks. For a non-linear 3DMM, a neural network or other non-linear model is adopted to model the mapping~\cite{bagautdinov2018modeling, tewari2018self, tran2019learning, tran2018nonlinear, cheng2019meshgan, tran2019towards}, which is more powerful in representing detailed shape and appearance than linear mapping.
3DMM is widely used in 3D face reconstruction~\cite{zhu2017face, zhu2016face, xiao2022detailed}, generation~\cite{zhuang2022mofanerf, hong2022headnerf, wu2023high}, talking faces~\cite{yu2022migrating, sun2023vividtalk, He2022Emo3DTalk}, and other downstream applications. The main problem with 3DMM is that it cannot model hair, beard, and eyelashes, since 3D polygon mesh models can hardly represent these structures. Therefore, traditional 3DMM cannot be used to render highly realistic images. This problem is solved in methods 2D generative model and conditional NeRF by introducing photometric loss and differentiable rendering. We recommend referring to the recent survey\cite{egger20203d} for a comprehensive review of 3DMM. By contrast,

\noindent\textbf{2D Generative Head Model: }
Research on deep generative models has made significant progress in the last decade. The conditional generative models represented by the generative adversarial network (GAN)~\cite{goodfellow2020generative} have produced excellent results in the task of 2D face generation~\cite{karras2019style, karras2020analyzing, deng2020disentangled, deng2022gram, chan2021pi}. These methods learn a distribution of a large-scale 2D face image dataset through a deep generative network, which is trained by a min-max game between the generator network and the discriminator network. GAN-based models can not only synthesize 2D faces given a parameter but also obtain the parameters that fit a given 2D face by GAN inversion methods~\cite{xia2022gan}.  As GAN-based models are commonly trained on a large amount of real-captured portraits, their generated faces are highly photo-realistic. The disadvantage of GAN-based models is that they lack 3D information, so large-angle view synthesis is not supported. We highly recommend reading the survey on GANs~\cite{creswell2018generative, gui2021review} and GANs for face~\cite{kammoun2022generative} for a comprehensive understanding of this field.  Very recently, diffusion models~\cite{ho2020denoising} for face generation have shown appealing performance in cross-modal face generation~\cite{kim2023dcface, huang2023collaborative}. However, similar to GAN-based generative models, they suffer from the lack of stable 3D information as well. 

\noindent\textbf{Conditional Head NeRF: }
NeRF~\cite{mildenhall2020nerf, mildenhall2021nerf} provides an alternative representation for the parametric head by combining an implicit neural network with a volume renderer. 
A straightforward idea is to combine NeRF with GAN to create a conditional 3D GAN~\cite{toshpulatov2021generative}. Early works~\cite{niemeyer2021giraffe, gu2021stylenerf, chan2022efficient, sun2023next3d, sun2022fenerf} train generative 3D GANs by using massive 2D real portraits and are surprised to find that the model can learn 3D information from these 2D portraits with various identities and views. However, the learned 3D information is limited and can be rendered only at small angles in the front, while larger angles will lead to severe rendering degradation. HeadNeRF~\cite{hong2022headnerf} proposes to train the model with both large-scale 2D head images and relatively few 3D models, but only extending the renderable view to about $60^\circ$. MoFaNeRF~\cite{zhuang2022mofanerf} only leverages high-quality multi-view head images for training and realizes a 180-degree free-view renderable parameter head model. However, the main problem is that existing high-quality multi-view 3D head data are commonly captured with hair tied up, so MoFaNeRF does not model hair. In later studies, RODIN~\cite{wang2023rodin} trains the parametric head with large-scale multi-view images rendered from artist-designed 3D models, and PanoHead~\cite{an2023panohead} combined large-scale frontal 2D images with captured back-view hair images for training, both of which achieved a $360^\circ$ renderable parametric head model. The main problem with these two models is that their facial expressions and motions are not riggable. Unlike all the above models, our model is the first motion-riggable and $360^\circ$ renderable high-fidelity parametric head model.

\section{Method}
\label{sec:method}

\begin{figure*}[th]
    \centering
    \includegraphics[width=1.0\linewidth]{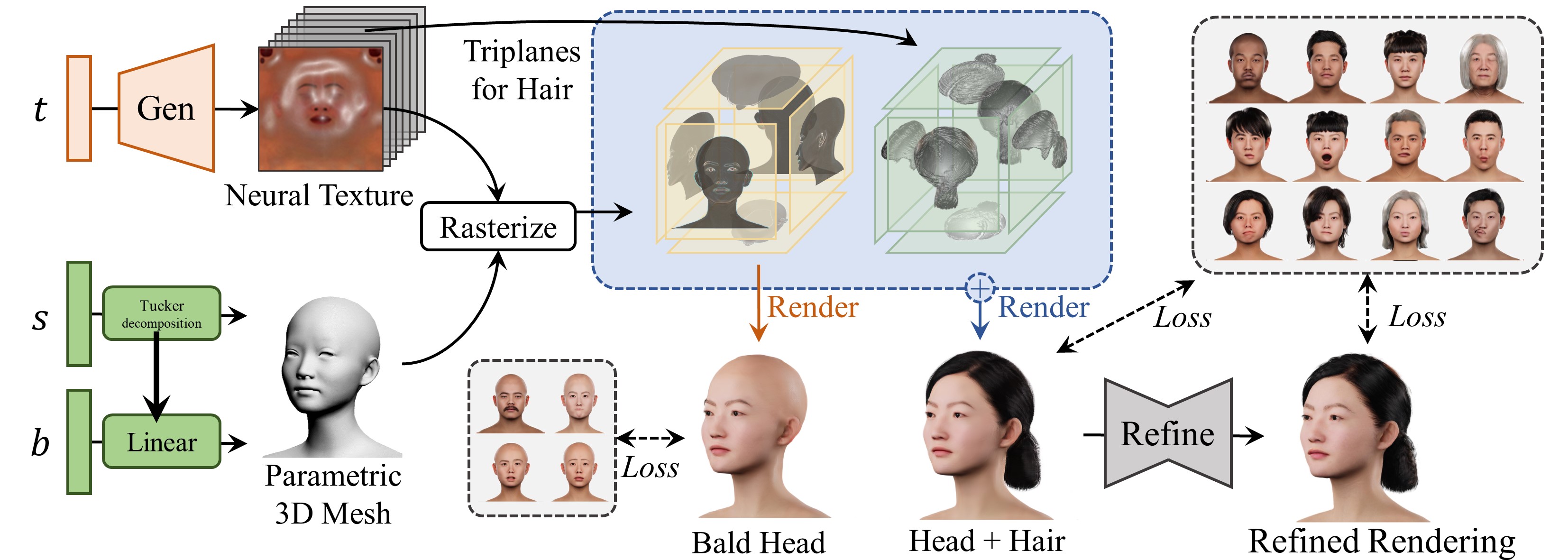}
    \caption{\textbf{Overall pipeline}. Our model is represented by a neural radiance field with hex-planes, conditioned on a generative neural texture and a parametric 3D mesh model. In this way, the facial appearance, shape, and motion are parameterized as texture code $t$, shape code $s$, and blendshapes parameter $b$, respectively. The RefineNet, a conditional GAN, is introduced to further improve the details of the generated faces.
    }
    \label{fig:pipeline}
\end{figure*}

We present a parametric 3D head model with hair for free-view synthesis in 360$^\circ$ and achieve single-image fitting, animating, and text-based editing. In this section, we will first introduce the artist-designed 3D head datasets in Sec.~\ref{sec:dataset}, which is used to train the parametric head. Then, the representation of the 3D head and the network design will be explained in Sec.~\ref{sec:rep}. Finally, the training methods and the model manipulations, including fitting, animation, and editing, will be presented in Sec.~\ref{sec:train} and Sec.~\ref{sec:mani}, respectively.

\subsection{\datasetname Dataset}
\label{sec:dataset}

The quality and quantity of the training data are crucial to the performance of a learning-based parametric head, but capturing ideal training data is very hard and expensive. As shown in Tab.~\ref{tab:data}, previous multi-view or 3D head datasets have certain obvious flaws, like low 3D accuracy~\cite{cao2013facewarehouse}, wrapped hair~\cite{yang2020facescape, dai2020statistical, wang2022faceverse, zhu2023facescape}, and lack of rigging~\cite{dai2020statistical, pan2023renderme}. Very recently, learning from large-scale synthetic 3D heads~\cite{wood2021fake} has proved to be effective~\cite{wang2023rodin}. However, the synthetic 3D head dataset they used is not publicly released and is a very high cost.

\begin{figure*}[th]
    \centering
    \includegraphics[width=1.0\linewidth]{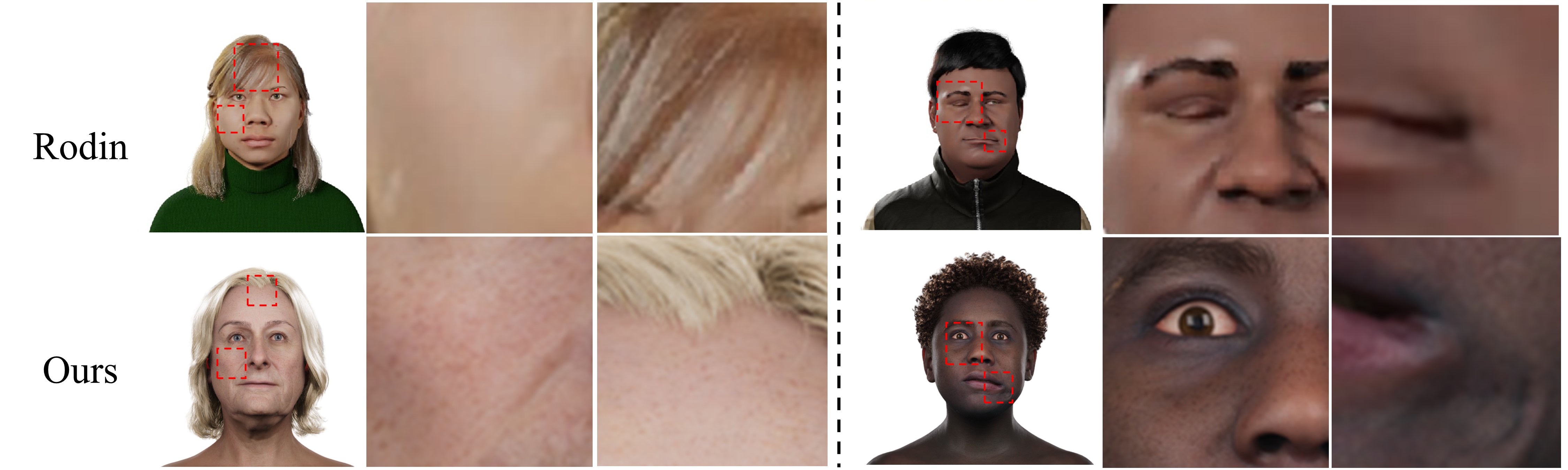}
    \caption{\textbf{Comparison of the data quality.}. Our images (\datasetname) exhibit a greater level of detail than Rodin~\cite{wang2023rodin}, including pores, wrinkles, and subtle textures. 
}
    \label{fig:with_rodin}
\end{figure*}

Therefore, we create a high-fidelity synthetic 3D head dataset for free research use, containing $100$ different subjects with diverse hairstyles and facial appearances, rigged into $52$ standard blendshapes bases. The models were designed by the artists referring to FaceScape~\cite{zhu2023facescape} and HeadSpace~\cite{dai2020statistical}, two public data sets, which cover the appearance of western and eastern races. 
The \datasetname dataset encompasses $374,400$ calibrated high-resolution images and $5,200$ mesh models for each identity under 52 expressions. The ratio of males to females in the dataset is 1:1, and the age is fairly evenly distributed between 16 and 70. The 3D heads are rendered by 72 head-centric virtual cameras covering 3 pitch angles and 24 horizontal rotation angles. Both rendered images and 3D mesh models will be released. Regarding model quality, the models of \datasetname dataset are more detailed and realistic than that of the Rodin dataset, as shown in Figure~\ref{fig:with_rodin}. Our models exhibit a greater level of skin detail, including pores, subtle textures, and the natural variations of skin irregularities. The rendering of light and shadow on facial features is more refined, reflecting a complex lighting environment that includes specular reflections off the skin's surface and soft shadow edges. More descriptions can be found in the supplementary material.

\begin{table}[]
\centering
\caption{\textbf{Comparisons of 3D head datasets.} 
}
\scalebox{0.90}{
\begin{threeparttable}
\begin{tabular}{@{}lcccccllccccc@{}}
\cmidrule(r){1-6} \cmidrule(l){8-13}
\multicolumn{1}{c}{Dataset}                                & Sub. & Range & Hair                       & Rig                        & \multicolumn{1}{l}{Source} &   & \multicolumn{1}{c}{Dataset}                               & \multicolumn{1}{c}{Sub.} & \multicolumn{1}{c}{Range} & \multicolumn{1}{c}{Hair}           & \multicolumn{1}{c}{Rig}    & Source                      \\ \cmidrule(r){1-6} \cmidrule(l){8-13} 
BU-3DFE~\cite{yin20063d}             & 100  & front & \ding{55} & \ding{51} & Active                     &  & BP4D-S~\cite{zhang2014bp4d}         & 41                       & \multicolumn{1}{c}{front} & \ding{51}         & \ding{51} & \multicolumn{1}{c}{Passive} \\
BU-4DFE~\cite{zhang2013high}         & 101  & front & \ding{51} & \ding{55} & Active                     &  & HeadSpace~\cite{dai2020statistical} & 1519                     & $270^{\circ}$             & \ding{55}         & \ding{55} & Active                      \\
BJUT-3D~\cite{baocai2009bjut}        & 500  & front & \ding{55} & \ding{55} & Active                     &  & FaceScape~\cite{zhu2023facescape}   & 938                      & $360^{\circ}$             & \ding{55}         & \ding{51} & Passive                     \\
Bosphorus~\cite{savran2008bosphorus} & 105  & front & \ding{55} & \ding{51} & Active                     &  & FaceVerse$^\dagger$~\cite{wang2022faceverse}  & 128                      & $360^{\circ}$             & \ding{51} & \ding{51} & Hybrid                      \\
FaceWarhouse\cite{cao2013facewarehouse}       & 150  & front & \ding{51} & \ding{51} & Active                     &  & Rodin$^\ast$~\cite{wang2023rodin}   & $10^5$            & $360^{\circ}$             & \ding{51}         & \ding{55} & Manual                      \\
4DFAB~\cite{cheng20184dfab}          & 180  & front & \ding{55} & \ding{55} & Hybrid                     &  & RenderMe360$^\dagger$~\cite{pan2023renderme}  & 500                      & $360^{\circ}$             & \ding{51} & \ding{55} & Passive                     \\
D3DFACS~\cite{cosker2011facs}        & 10   & front & \ding{51} & \ding{51} & Passive                    &  & SynHead100(Ours)                        & 100                      & $360^{\circ}$             & \ding{51}         & \ding{51} & Manual                      \\ \cmidrule(r){1-6} \cmidrule(lr){8-13}

\end{tabular}

 \begin{tablenotes}
        \footnotesize
        \item[$\ast$] Dataset not publicly available; \quad $\dagger$  Rough captured 3D hair shape.
      \end{tablenotes}
    \end{threeparttable}
}
\label{tab:data}
\end{table}

\subsection{Head Representation}
\label{sec:rep}

As shown in Fig.~\ref{fig:pipeline}, our model is represented by a neural radiance field~\cite{mildenhall2020nerf} with hex-planes, which is conditioned on a generative neural texture~\cite{sun2023next3d, thies2019deferred} and a parametric 3D mesh model~\cite{li2017learning}. 

\noindent\textbf{Parametric 3D Mesh.} 
Given the shape code $s$ and blendshapes code $b$ representing facial motion, a head avatar mesh can be generated by a mapping defined by the parametric 3D mesh model. This mapping is established by reducing dimensionality to the synthetic dataset on the identity dimension. Specifically, given the vertices of the mesh models in our synthetic dataset containing 100 identities $\times$ 52 blendshapes bases, we first create a large matrix $V_o \in \mathbb{R}^{(3N) \times 100 \times 52}$, where $N$ denotes the number of the vertices. Then tucker decomposition~\cite{tucker} is used on the identity dimension, yielding a smaller core tensor $C_r \in \mathbb{R}^{(3N)\times50\times52}$, the bilinear model face model for our model. In this way, the mapping between 3D models and parameters ($s$ for facial shape and $b$ for facial expressions) can be formulated as:

\begin{equation}\label{eq:core}
    V = C_r \times s \times b
\end{equation}

\noindent where $V$ is a $N \times 3$ vector representing vertices of a head avatar model. 

\noindent\textbf{Dual Hex-planes for Detaching Hair.} 
Given the texture code $t$, the neural texture and the feature maps are synthesized by an image generator network. Following Next3D~\cite{sun2023next3d}, the neural textures with the generated 3D mesh are rasterized into multiple feature planes and rendered to the image given the camera parameters. Finally, RefineNet further improves the details of the rendered images. 
Based on the representation of tri-planes NeRF conditioned on neural texture and 3D mesh, we further propose three modifications.
First, instead of three feature planes to be rasterized~\cite{chan2022efficient}, we adopt six feature planes, namely hex-planes, to model the $360^\circ$ appearance of the head from front, back, left, right, top, and bottom. Experiments showed that the hex-planes improve the rendering quality and are more suitable for training on $360^\circ$ multi-view image data. Second, A two-branch network is introduced to model a bald head and hair separately. We propose a training strategy that decouples hair and head to achieve free-swapping of hairstyles and better fitting performance. The training strategy for disentangling hair and head will be detailed in Sec.~\ref{sec:train}.  Finally, a more accurate parametric 3D head model is used, which is connected to the rasterizing module. Though the parametric head model is not trained with neural texture, the connection enables the shape code $s$ to be an optimizable parameter in the GAN inversion.

\subsection{Model Training}
\label{sec:train}

\noindent\textbf{Loss Functions.}
The loss function to train the hex-planes conditioned on neural texture and parametric 3D mesh is formulated as: 

\begin{equation}
    \begin{split}
        \mathcal{L}_{total} = \mathcal{L}_{photo} + \lambda_{1}( \mathcal{L}_{d-gan} + \lambda_{2} \mathcal{L}_{density})
    \end{split}
 \end{equation}

\noindent where $\mathcal{L}_{photo}$ is the photometric loss, which is calculated by $L_1$ loss between the renderings and the ground-truth image; $\mathcal{L}_{d-gan}$ is the dual GAN loss presented in Next3D~\cite{sun2023next3d}; $\mathcal{L}_{density}$ is the density regularization proposed in EG3D~\cite{chan2022efficient}. In our experiments, the loss weights $\lambda_{1}$ is set to $0.01$, and $\lambda_{1}$ is set following the setting of EG3D. The learning rate $l$ is related to the regularization interval $R_i$ of the generator and discriminator, which is formulated:

\begin{equation}
    \begin{split}
        l &= l_{base} \times \frac{R_i}{Ri + 1}
    \end{split}
\end{equation}

\noindent where the $R_i$ for the generator is $4$ and that for the discriminator is $16$ in our experiments; Adam optimizer is used to train the model with the base learning rate $l_{base}$ as $0.0025$ and the batch size as $8$.

\noindent\textbf{RefineNet.}
A conditional GAN is introduced to further improve the details of the generated faces. Following MoFaNeRF~\cite{zhuang2022mofanerf}, we adopt pix2pixHD~\cite{wang2018high} as the backbone of our RefineNet. The input of the RefineNet is the generated images, which are rendered from the composited hex-planes. In the training phase, the overall network except for RefineNet is first trained, then RefineNet is trained with the other parts of the network detached. The loss function and hyper-parameters for training RefineNet are the same as pix2pixHD~\cite{wang2018high}.

\noindent\textbf{Implementation Details.}
Our model is trained on 4 NVIDIA 3090 GPUs for roughly 10 days. To achieve a balance between training speed and rendering quality, the images in our dataset are downscaled to a resolution of $512\times512$ before being fed into the network, and the rendering structure of the model also outputs images at a resolution of $512\times512$. 
As introduced in Sec.~\ref{sec:rep}, dual hex-planes are generated for head and hair separately. The full network with two hex-planes trainable is firstly trained with the images with hairs in $3000$k iterations, then is fine-tuned for hair detachment for additional $40$k iterations.
Due to limited space, we put more training details in the supplementary material.

\subsection{Model Manipulation}
\label{sec:mani}

\noindent \textbf{Fitting to Single Image.} 
A novel method is proposed to fit our model to a target image. Initially, we followed GAN-inversion~\cite{xia2022gan} to keep the trained parameters of the model unchanged and optimize for shape code $s$ and texture code $t$, but found the fitting results bad.  Therefore, three improvements are made to improve the fitting performance according to our observations and experiments.

Firstly, we believe that one major reason is that the solution space formed by $s$ and $t$ is too large to be optimized simultaneously.  We also tried to optimize $t$ and $s$ iteratively, again with no good results. Therefore, we propose to obtain a relatively faithful shape code $s$ by minimizing the projection error of facial landmarks, according to Eq. ~\ref{eq:core} as presented by Yang~\etal~\cite{yang2020facescape, zhu2023facescape}. Then, the shape code is set unchanged, and only texture is optimized via GAN-inversion. 

Secondly, considering that our parametric head is trained on relatively small data with only 100 identities, the parametric space of the texture code $t$ is very limited to cover diverse facial appearances. To address this issue, we propose to optimize for the Neural Texture rather than texture code $t$ or mapped texture code $w$, which further boosts the generalization of the fitting. Specifically, L1 loss between the generated frontal views and the input image is adopted for the single-image fitting. The shape code $s$ is optimized to obtain a fitted mesh that matches the input image. The fitted shape is then fixed and a random neural texture is selected as the initialization to optimize the neural texture. The image is firstly normalized to align the input face and the canonical position of our parametric 3D head. Then, Poisson blending~\cite{perez2023poisson} is used on the output image obtained from random texture initialization with wild real faces that have been aligned to \datasetname dataset. By optimizing the hex-planes space to reduce the discrepancy between the pre-processed images and those generated with random textures, the neural texture and latent code for hair can be optimized. Then, a $360^\circ$ renderable and animatable head is generated to fit the appearance of the input image.

Thirdly, despite implementing the aforementioned improvements, we noticed that the fitting results for hairstyles remain unsatisfactory. We believe this is due to the extensive diversity in everyday hairstyles, which our dataset of 100 hairstyle sets is inadequate to represent. To alleviate this problem, we propose to first remove the hair from the target images by semantic parsing algorithm~\cite{yu2018bisenet, yu2021bisenet} and optimize for a bald head model. Then a hairstyle classifier based on ResNet-50~\cite{he2016deep} is trained to classify the target's hairstyle into $30$ categories that match the hairstyle categories in our \datasetname dataset.  The model is trained by $6200$ face images synthesized by StyleGanV2 with manual annotations of hairstyle. Finally, the predicted hairstyle is combined with the fitted bald head model to generate a high-quality head model with hair.

\noindent \textbf{Animation.} Our parametric 3D mesh is defined following a standard $52$ facial blendshapes, which is also adopted by ARKit~\cite{arkit}. Therefore, our generated or fitted head can be animated by many blendshapes streaming tools. Our model can synthesize appearance features not closely attached to the mesh surface (like hair, mouth, and eyelashes), which are temporal consistent. It means that our proposed learnable representation can reach the performance of complex traditional artist-designed head avatars, achieving high-fidelity animation. The animation results are shown in Sec.~\ref{sec:exp} and the supplementary video.

\noindent \textbf{Text-based Editing.} 
Inspired by Instruct-NeRF2NeRF~\cite{instructnerf2023}, we incorporate text-based image-conditioned diffusion model~\cite{Brooks_2023_CVPR} with our animatable 3D parametric representation for 3D head editing. 
Specifically, after a head is generated or fitted, the multi-view images are rendered and edited by a text-guided image-to-image translator. The edited images are then used to fine-tune our parametric 3D head model. Empirically, we sample patches on generated images to fine-tune our parametric model, leading to better performance than the original random sampling strategy. Our text-based editing strategy aims to add abstract features, such as `wearing makeup' and `aged', to a generated head, without changing his/her identity.
The edited head can still be animated by standard blendshapes parameters.

\begin{figure*}[tb]
    \centering
    \includegraphics[width=0.97\linewidth]{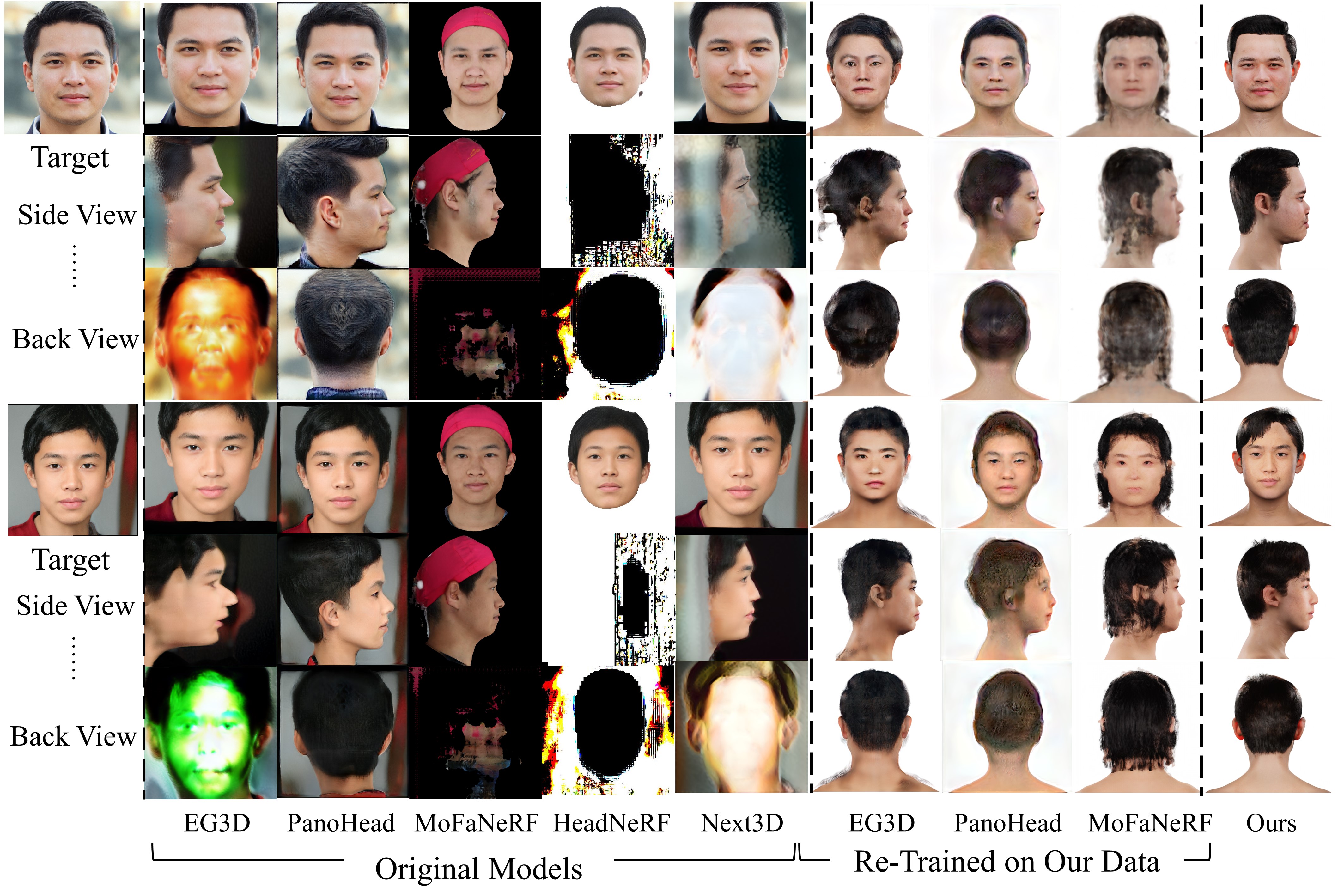}
    \caption{
    \textbf{Comparison of fitting results.} We compare our method with previous parametric or generative 3D head models in single-image fitting. For a comprehensive comparison, both original models and re-trained models are compared.
    }
    \label{fig:comp_fit}
\end{figure*}

\begin{table}[tb]

\centering
\caption{\textbf{Quantitive evaluation of fitting results}. 
}
\begin{tabular}{@{}lcccc@{}}
\toprule
Method     & PSNR(dB)↑  & SSIM↑($\times0.1$)       & LPIPS↓($\times0.1$)  & CSIM↑(×0.1)     \\ \midrule
EG3D~\cite{chan2022efficient}    & 13.25±0.79 & 5.97±0.10 & 3.27±0.12 &8.09±0.08\\
MoFaNeRF~\cite{zhuang2022mofanerf}  & 15.74±0.20 & 7.72±0.10 & 2.08±0.12&4.99±0.13 \\
PanoHead~\cite{an2023panohead}  & 15.10±1.68 & 6.07±0.22 & 2.42±0.05&7.60±0.11 \\
HeadNeRF~\cite{hong2022headnerf} & 10.44±0.54 & 6.86±0.07 &3.30±0.02&4.93±0.01 \\
Nextd3D~\cite{sun2023next3d}  &  15.72±2.90 & 6.43±0.52 & 2.24±0.71&7.81±0.09\\
EG3D(re-trained)~\cite{chan2022efficient}  &16.68±0.72&7.81±0.04 &1.79±0.21&4.79±0.19\\
MoFaNeRF(re-trained)~\cite{zhuang2022mofanerf}  & 13.31±0.27  &7.42±0.05   &2.83±0.25&3.92±0.05\\
PanoHead(re-trained)~\cite{an2023panohead}  & 17.18±1.5&7.92±0.03  & 1.89±0.18&6.89±0.15 \\
Ours & \textbf{17.41±0.13} &  \textbf{8.73±0.05} & \textbf{1.18±0.11}&\textbf{8.74±0.09}  
\\\bottomrule
\end{tabular}
\label{tab:fit}
\end{table}

\section{Experiments}
\label{sec:exp}

In this section, we first compare our fitting results and generated results with previous methods, then analyze the effectiveness of the proposed module through the ablation study, and finally show the results of animation, hairstyle-swapping, and text-based editing.

\subsection{Comparison of Fitting Results}

We compare our method with state-of-the-art parametric or generative 3D head models in the task of single image fitting, including EG3D~\cite{chan2022efficient}, PanoHead~\cite{an2023panohead}, HeadNeRF~\cite{hong2022headnerf}, MoFaNeRF~\cite{zhuang2022mofanerf}, and Next3D~\cite{sun2023next3d}. Both original and re-trained models on our dataset are compared for a comprehensive comparison.

All the methods are compared quantitatively in Tab.~\ref{tab:fit} and qualitatively in Fig.~\ref{fig:comp_fit}. 
Regarding original models, EG3D and Next3D are originally trained on large-scale in-the-wild 2D face datasets. The fitted head is of high quality at the frontal views but fails when viewing angles are larger than $90^\circ$. The PanoHead model is trained on both large-scale in-the-wild 2D images and in-house captured hair and head images. PanoHead synthesizes plausible results in $360^\circ$, but the resulting model cannot be animated like ours. The MoFaNeRF model is trained on studio-captured high-quality 3D face models. Its results are canonical heads without hair and fail to be rendered when the viewing angles are larger than $90^\circ$ as well. HeadNeRF leverages both in-the-wild images and studio-captured multi-view images for training. Its results are animatable but degrade severely when viewing angles are larger. In summary, the fitting results of all these methods are plausible at frontal views, but cannot be rendered in $360^\circ$ except for PanoHead. The major drawback of PanoHead is that its results are static and cannot be animated.

Regarding re-trained models on our dataset, we evaluate the performance of re-trained EG3D, PanoHead, and MoFaNeRF, as shown in Fig.~\ref{fig:comp_fit} and Tab.~\ref{tab:fit}. The fitting method of all these three models is to optimize for the latent code by GAN inversion strategy. Since the appearances in our training set are few ($100$ identities), the performance of retrained EG3D, PanoHead, and MoFaNeRF is poor due to the limited appearance space. By contrast, our method optimizes the neural texture to fit the input image, which enables the model to fit in a much larger solution space. The visual comparison shows that our results are more plausible in $360^\circ$ than all the re-trained models, which is also verified by the quantitative evaluations.
\begin{table}[t]
\centering
\caption{\textbf{Quantitative evaluation of generated results} }
\begin{tabular}{@{}lccc@{}}
\toprule
Method     & PSNR(dB)↑  & SSIM↑($\times0.1$)       & LPIPS↓($\times0.01$)        \\ \midrule
EG3D~\cite{chan2022efficient}      & 26.45±3.27 & 8.50±0.45 & 6.60±2.24 \\
MoFaNeRF~\cite{zhuang2022mofanerf}  & 28.09±2.36 & 8.60±0.32 & 7.45±1.73 \\
PanoHead~\cite{an2023panohead}  & 25.89±3.88 & 8.60±0.46 & 10.26±2.83 \\
Vanilla NeRF & 24.44±2.65 & 8.26±0.37 &8.16±1.68 \\
Tri-plane&  25.66±2.14 &  8.27±0.13 &  9.54±1.72\\
Ours - no refine      & \textbf{28.92±2.32} & \textbf{8.83±0.01} & 7.31±1.75 \\
Ours - full & 28.27±2.61 & 8.66±1.42 & \textbf{6.31±1.71} \\
\bottomrule
\end{tabular}
\label{tab:comp}
\end{table}

\begin{figure}[t]
    \centering
    \includegraphics[width=1.0\linewidth]{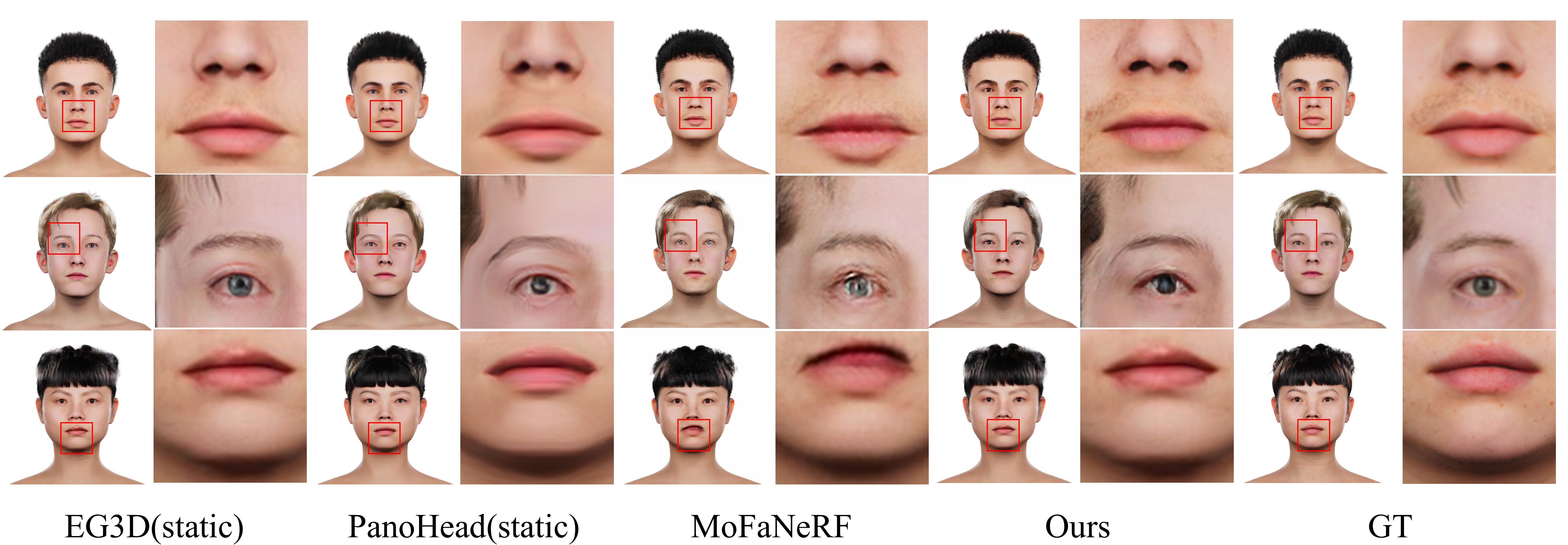}
    \caption{\textbf{Comparison of generated results.}
    }
    \label{fig:comp_gen}
\end{figure}

\begin{figure}[th]
    \centering
    \includegraphics[width=1.0\linewidth]{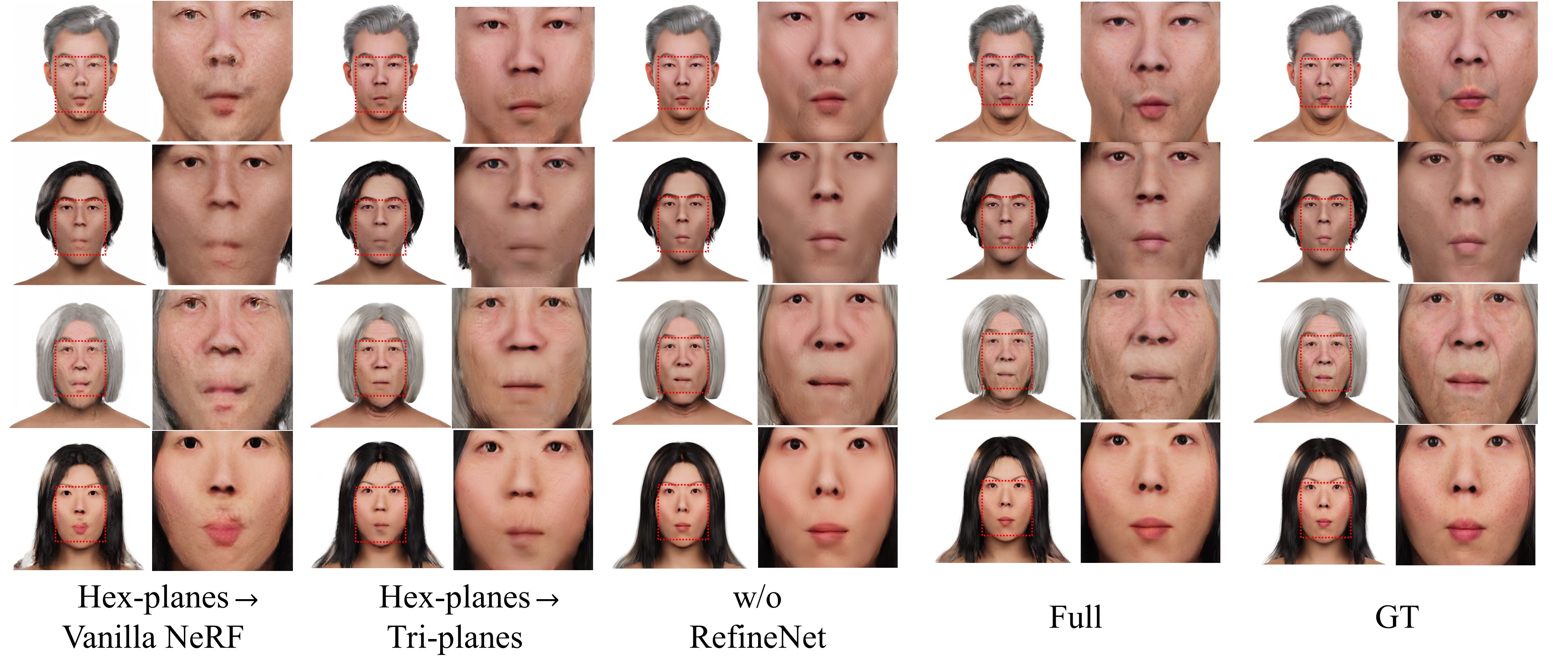}
    \caption{\textbf{Ablation study.} 
    The performance of our full model outperforms all the ablated settings with more clear and detailed rendering, which proves the effectiveness of these proposed modules. The issues are highlighted with the red dotted box. 
    }
    \label{fig:ablation}
\end{figure}

\begin{figure}[th]
    \centering
    \includegraphics[width=1.0\linewidth]{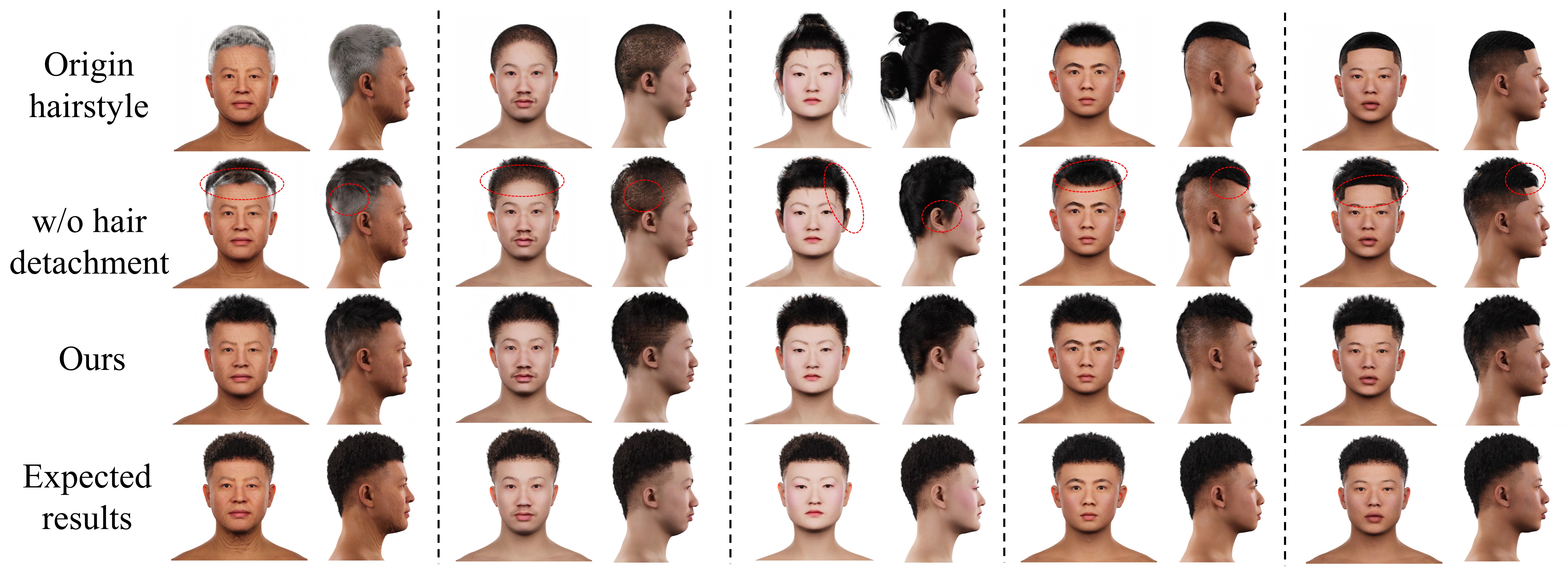}
    \caption{\textbf{Ablation study about hair detachment.} 
    Given a generated or fitted head, the original hairstyle is replaced with a certain hairstyle by two models.  
    The face-swapping results with hair detachment are more plausible with fewer artifacts.
    }
    \label{fig:hair_ablation}
\end{figure}

\begin{table}[t]
\centering
\caption{\textbf{Quantitive Evaluation of Hair Swapping}. 
}
\begin{tabular}{@{}lccc@{}}
\toprule
Label     & PSNR(dB)↑  & SSIM↑($\times0.1$)       & LPIPS↓($\times0.01$)        \\ \midrule
w/o hair detachment      & 20.04±3.83  & 8.35±0.24 & 15.89±2.36 \\
Ours  & \textbf{21.86±2.22} &\textbf{ 8.40±0.22} & \textbf{10.51±2.01} \\\bottomrule
\end{tabular}
\label{tab:hair}
\end{table}

\begin{figure}[th]
    \centering
    \includegraphics[width=1.0\linewidth]{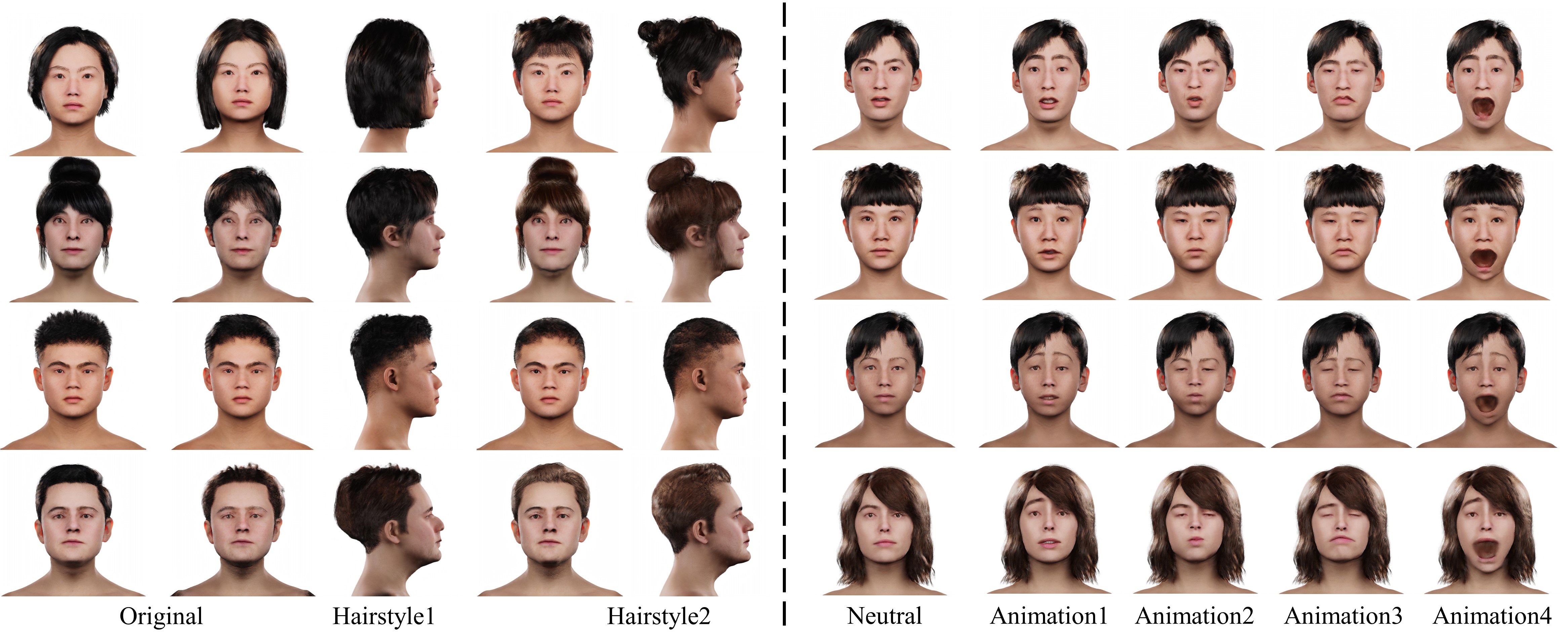}
    \caption{\textbf{Hair swapping and animated results.} Left: Given a generated head, the original hairstyle can be replaced with other hairstyles. Right: our generated head can be driven by blendshapes parameters, with facial details accurately synthesized. 
    }
    \label{fig:hair_and_anim}
\end{figure}

\begin{figure}[th]
    \centering
    \includegraphics[width=1.0\linewidth]{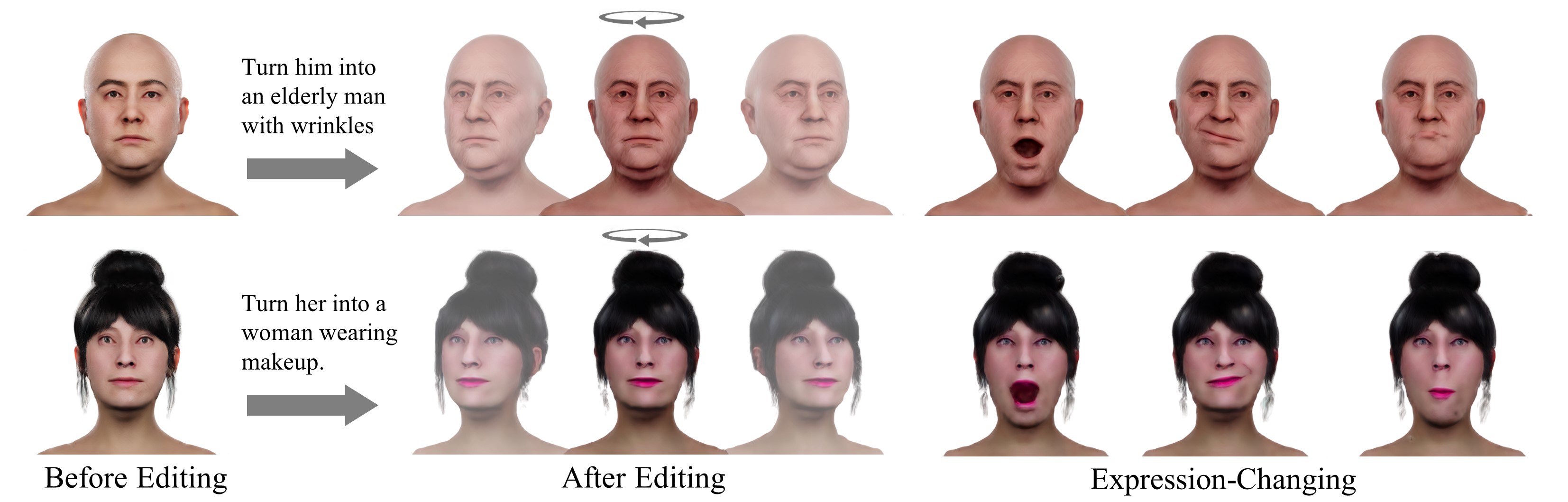}
    \caption{\textbf{Text-based editing results.} Given a text-based input (italics sentences), our generated model can be further edited to match the prompt while keeping identities unchanged. The animated frames of the edited head are shown on the right.
    }
    \label{fig:text}
\end{figure}

\subsection{Comparison of Generated Results}

We compare our method with several pervious parametric or generative 3D head models, including EG3D~\cite{chan2022efficient}, PanoHead~\cite{an2023panohead}, HeadNeRF~\cite{hong2022headnerf}, and MoFaNeRF~\cite{zhuang2022mofanerf}.
The original models in their papers are trained on large-scale 2D image datasets or reconstructed 3D datasets, which is quite different from the synthetic dataset we used. For a fair comparison, we re-train EG3D, PanoHead, and MoFaNeRF on our synthetic dataset. The generated faces of the same identity in the training set are compared to evaluate the representation ability of these models.

The qualitative comparison is shown in Fig.~\ref{fig:comp_gen}. The aim of EG3D and PanoHead is modeling static 3D models that cannot be animated, which is relatively easier. The results of EG3D, PanoHead, and ours are of comparable quality, while our results are better in facial details. Our performance is also better than another animatable parametric 3D head, MoFaNeRF. 

The quantitative comparison is shown in Tab.~\ref{tab:comp}, where PSNR~\cite{hore2010image}, SSIM~\cite{wang2004image}, LPIPS~\cite{zhang2018unreasonable}, and CSIM~\cite{csim} are for method are reported. For a fair comparison, all the models to be evaluated are re-trained on our artist-designed dataset. EG3D and PanoHead cannot be animated as the expression is not parameterized in their model, so only neutral expressions are used for evaluation. By contrast, the results of MoFaNeRF and our model are riggable.  Our method outperforms previous methods in all four metrics, which demonstrates the effectiveness of our representation in $360^\circ$ head modeling.

\subsection{Ablation Study}
We perform ablation studies on specific modules as follows:
\begin{itemize}
    \item \textit{Hex-planes $\longrightarrow$ Vanilla  NeRF}. The neural texture and hex-planes (Sec.~\ref{sec:rep}) are replaced by a conditioned vanilla NeRF.
    \item \textit{Hex-planes $\longrightarrow$ Tri-planes}. The hex-planes representation (Sec.~\ref{sec:rep}) is replaced by a classic tri-planes representation.
    \item \textit{w/o RefineNet}. The RefineNet (Sec.~\ref{sec:train}) is removed.
\end{itemize}

As reported in Tab.~\ref{tab:comp}, Fig.~\ref{fig:ablation}, adding RefineNet slightly decreases PSNR and SSIM scores, while enhancing LPIPS score. We think the reason is that RefineNet synthesizes photo-realistic details that conform to perceptual features (higher LPIPS). Though these details do not match the ground truth details (lower PSNR and SSIM), synthesizing such realistic details is visually better, as visualized in Fig.~\ref{fig:comp_gen}.
Additionally, ablation experiments are conducted to verify the effectiveness of the hair detaching module, as reported in Tab.~\ref{tab:hair} and Fig.~\ref{fig:hair_ablation}.

\subsection{Other Manipulations}

\noindent \textbf{Hair-Swapping.} As the hair and head are detached in the hidden feature space, our model supports free hair-swapping for a generated or fitted head. As shown on the right side of Fig.~\ref{fig:hair_and_anim}, given a generated head on the left, the original hairstyle can be replaced with others.

\noindent \textbf{3D Animation.}  The generated results can be driven by blendshapes parameter streams.
As shown in Fig.~\ref{fig:hair_and_anim}, the facial details like mouth, eyes, and motion wrinkles of a dynamic head are accurately modeled.

\noindent \textbf{Text-based Editing Results.} As shown in Fig.~\ref{fig:text}, our generated 3D model can be further edited given a text prompt. The subject's identity is maintained after the editing, while abstract features like `makeup' and `aged' are synthesized.

\section{Conclusion}
\label{sec:con}
In this paper, we have constructed a dataset of artist-designed, high-fidelity human heads, and developed a novel framework to learn a $360^\circ$ free-view renderable parametric model from this dataset. Our approach decouples the facial motion/shape and facial appearance, represented by a classic blendshapes model and neural texture, respectively. Notably, our model is the first of its kind — a parametric 3D head model that supports $360^\circ$ free-view synthesis, single-image fitting, and animation driven by a blendshapes parameter flow.

\noindent \textbf{Limitations.} 
Creating high-fidelity avatars comes at a considerable cost, thus limiting our dataset to 100 identities.
Considering the high cost of producing high-fidelity digital human data, we think it is meaningful to study how to learn a parametric head model under the condition of limited data amount.
Although some in-the-wild fitting results are plausible by learning from this finite training set, stability in fitting remains a challenge. Evidence of this can be found in the failure case presented in our supplementary material.  A potential solution might entail expanding the dataset and incorporating real-world images and models into the training process.
Moreover, our method does not decompose the material and lighting, which results in some highlights being baked into the texture. It reduces the photo-realism and application potential. This issue could be alleviated by utilizing the illumination and material information from the synthetic dataset.
We leave these considerations for future work.

\noindent \textbf{Acknowledgement.} This study was funded by NKRDC 2022YFF0902200, NSFC 62441204, and Tencent Rhino-Bird Research Program.

%
%
\bibliographystyle{splncs04}
\bibliography{main}

\end{document}